\documentclass[letterpaper]{article}
\usepackage{natbib}
\usepackage{bm}
\usepackage{bbm}
\usepackage{graphicx}

\usepackage{amsmath}
\usepackage{amssymb}
\usepackage{tabularx}
\usepackage{booktabs}
\usepackage{makecell}
\usepackage{url,hyperref,cleveref}

\newcommand{\be}{\begin{eqnarray}}
\newcommand{\ee}{\end{eqnarray}}

\usepackage{alifeconf}
\newcommand\blfootnote[1]{%
  \begingroup
  \renewcommand\thefootnote{}\footnote{#1}%
  \addtocounter{footnote}{-1}%
  \endgroup
}

\title{Can AI Detect Life? Lessons from Artificial Life}
\author{Ankit Gupta$^{1,2,\star}$\and Christoph Adami$^{2,3,4,\dagger}$ \\
\mbox{}\\
$^1$Department of Computer Science and Engineering\\
$^2$Program in Evolution, Ecology, and Behavior \\
$^3$ Department of Microbiology, Genetics, and Immunology\\
$^4$ Department of Physics and Astronomy\\
Michigan State University, East Lansing, MI\\ 
$^\star$guptaa23@msu.edu,$^\dagger$adami@msu.edu}

\begin{document}
\maketitle

\begin{abstract}
Modern machine learning methods have been proposed to detect life in extraterrestrial samples, drawing on their ability to distinguish biotic from abiotic samples based on training models using natural and synthetic organic molecular mixtures. Here we show using Artificial Life that such methods are easily fooled into detecting life with near 100\% confidence even if the analyzed sample is not capable of life. This is due to modern machine learning methods' propensity to be easily fooled by out-of-distribution samples. Because extra-terrestrial samples are very likely out of the distribution provided by terrestrial biotic and abiotic samples, using AI methods for life detection is likely to yield significant false positives.
\end{abstract}
\vskip -0.5cm
\blfootnote{\textcopyright  2026 A. Gupta and C. Adami. Published under a Creative Commons Attribution 4.0 International (CC BY 4.0) license.}

\section{Introduction}
Creating an agnostic life detector (a system that detects life even in the absence of information about the biochemistry of life-bearing molecules) has been one of the  main goals of astrobiology for decades~\citep{Walkeretal2018}. Potential biosignatures can be classified into substances (e.g., isotopes, elements, and molecules), objects (e.g., fossils, stromatolites, etc.), and patterns (mineral, physical, or chemical)~\citep{Chanetal2019}. Which substances, patterns, or objects are diagnostic of life is, however, a difficult question since we are privy only to one form of life on this planet. Machine learning methods have been deployed in the past to classify amino acids as biotic or abiotic~\citep{Dornetal2003}, and agnostic biomarkers have been developed and even tested with Artificial Life~\citep{Dornetal2011,DornAdami2011}. However, the machine learning methods of today (dubbed ``AI" in the following) are far more powerful than the methods available in the past, and we can ask whether such methods are capable of generalizing from a sufficient number of natural and synthetic molecules to classify potentially ambiguous biomarkers. Indeed, recent efforts using AI to analyze data obtained from pyrolysis-gas chromatography-mass spectrometry (py-GC-MS)---a typical instrument that would be carried by dedicated space missions---suggested that AI can discriminate between abiotic and biotic samples with accuracies approaching 98\% accuracy~\citep{Cleavesetal2023,Hystadetal2025,Wongetal2025}. However, any claim of the discovery of a biosignature is likely to be greeted with skepticism, as the likelihood of false positives is generally high~\citep{SmithCole2023}.

Here we use Artificial Life to test whether an AI classifier (a multi-layer perceptron in  the focal experiments, and several other AI methods in the supplementary experiments) can be fooled into misclassifying a potential biomarker molecule that is a polymer built from a particular alphabet. Under the assumption that all life anywhere must encode information into unbranched heteropolymers~\citep{Adami2024}, testing whether AI can discriminate between living (self-replicating) and non-living (non-replicating) polymers can provide some insight into whether AI can detect life. Of course, weakening any of these assumptions weakens the conclusions we reach, but some of the lessons learned may apply beyond the particular case we are testing.

AI methods such as Deep Neural Networks (DNN)~\citep{Goodfellowetal2016} work by fitting a model to the distribution of samples within a training set, but exhibit significant vulnerability to out-of-distribution samples~\citep{Szegedyetal2013,Goodfellowetal2014}. For example, \cite{Nguyenetal2015} used evolution to find images that were confidently mis-classified by a DNN even though they were completely unrecognizable to the human eye. \cite{Guptaetal2025} used an even simpler procedure---a simple greedy hill climber from a black canvas---to produce images fooling the AI with only a few colored pixels splattered on a 224$\times$224 canvas. That work, fooling the most advanced AI classifiers on the largest dataset (ImageNet), showed that when the feature space is large enough, the number of out-of-distribution samples dwarfs the number of samples in the training set exponentially, so that arbitrary many fooling images are within the basins of attraction formed by the training procedure. 

It might seem impossible to test an AI's generalization power for life detection because out-of-distribution samples are usually not available for life. Artificial Life provides a unique opportunity to overturn this paradigm, because in some cases we have access to the {\em entirety} of the distribution, while presenting only a small subset of those to the AI. If the space of possible life forms (the feature space) in an Artificial Life system is small, however, it is unlikely that out-of-distribution samples are a significant subset of all possible samples. Fortunately, we have a dataset of significant size that may allow us to test the hypothesis that AI systems can be fooled into misclassifying life. The Digital Life system 
Avida~\citep{AdamiBrown1994,Adami1998,Ofriaetal2009} converts a computer's memory into a digital Petri dish for self-replicating computer programs that evolve to perform complex computations on numbers they are fed as ``food". Experiments using this digital life form have led to the discovery of a number of novel evolutionary processes that have subsequently been found in biological organisms~\citep{Adami2006b}.

\begin{figure}[htbp] 
   \centering
   \includegraphics[width=3in]{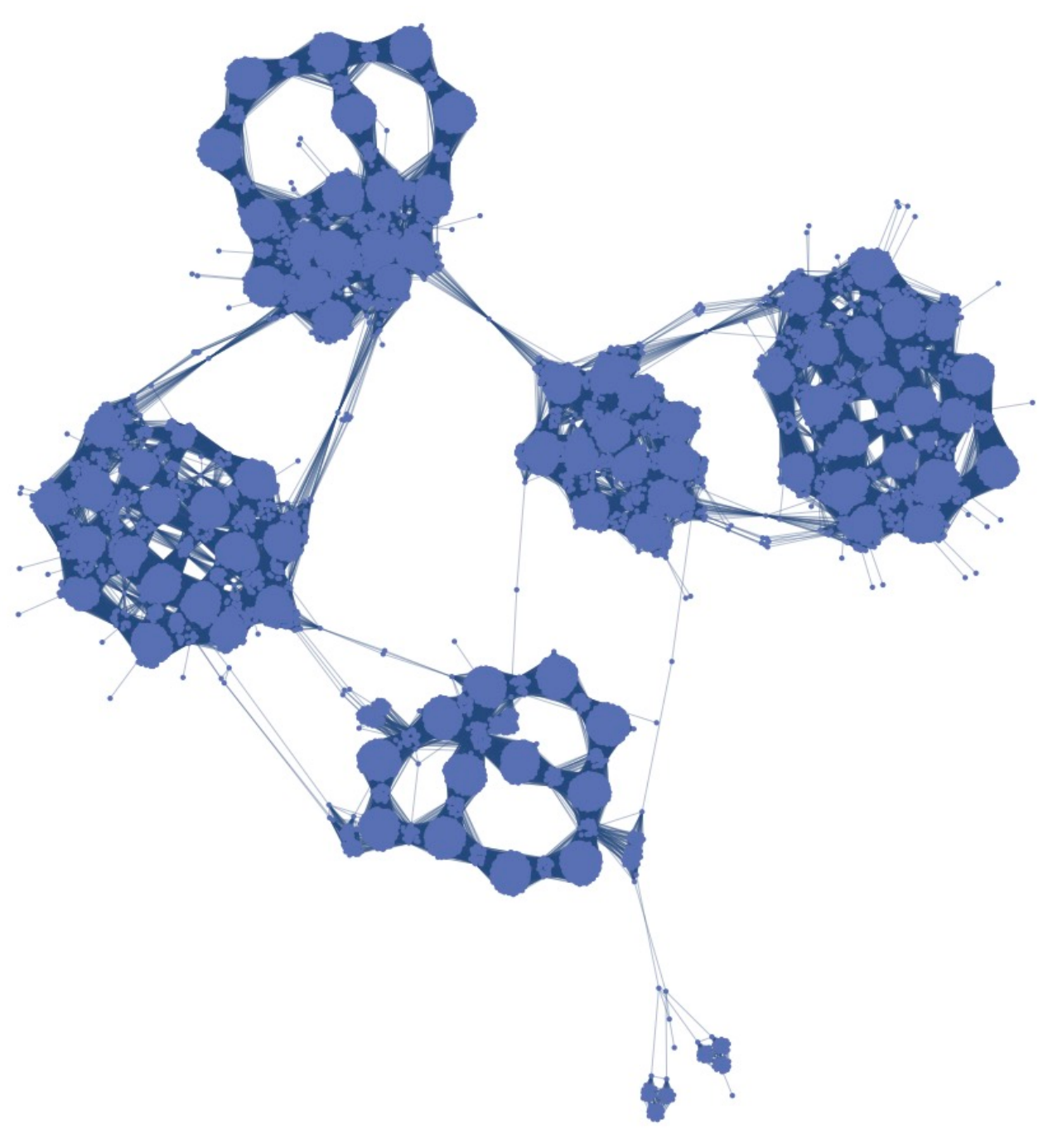} 
   \caption{Network representation of the largest cluster of avidian replicators of length 9, containing over 94\% of the replicators, with edges drawn between one-mutant neighbors. This graph consists of five sparsely connected groups of tightly-connected clusters. }
   \label{fig1}
\end{figure}

In a standard implementation, avidian programs are constructed from an alphabet of 26 different machine instructions running on a simulated CPU (to avoid contaminating real computers with evolvable malware). Because self-replicating programs are extremely rare in this language, ``ancestral" programs that are used to initiate evolution experiments are hand-written. An investigation into the likelihood that self-replicating programs could emerge via a chance process~\citep{AdamiLaBar2017} led to a project to map {\em all} possible small self-replicators. A search of all programs of length 6 and 7 found none. But an exhaustive search of all length 8 programs found 914 ``colony-forming" programs~\citep{Nitashetal2017} among the $26^8\approx209\times 10^{9}$ possible programs\footnote{Because self-replication does not require making exact copies, programs that create variants that are able to replicate so that a sustained growing group is formed are also deemed to be ``viable". Thus, our criterion for viability is the same as that for bacteria growing in a dish with nutrients.}. An even more extensive search of all $26^9\approx 5.43\times 10^{12}$ programs of length 9~\citep{NitashAdami2021} found exactly 36,171 viable sequences, which form clusters in sequence space (Fig.~\ref{fig1}).

The viable sequences occupy a fraction of 6.66$\times10^{-9}$ of the total feature space, corresponding to just a little under 6 mers of information (out of the 9 mers of sequence entropy). We will use this large fitness landscape to test whether we can fool modern AI to misclassify non-replicators as replicators with high confidence.

\begin{figure}[htbp] 
   \centering
   \includegraphics[width=2.5in]{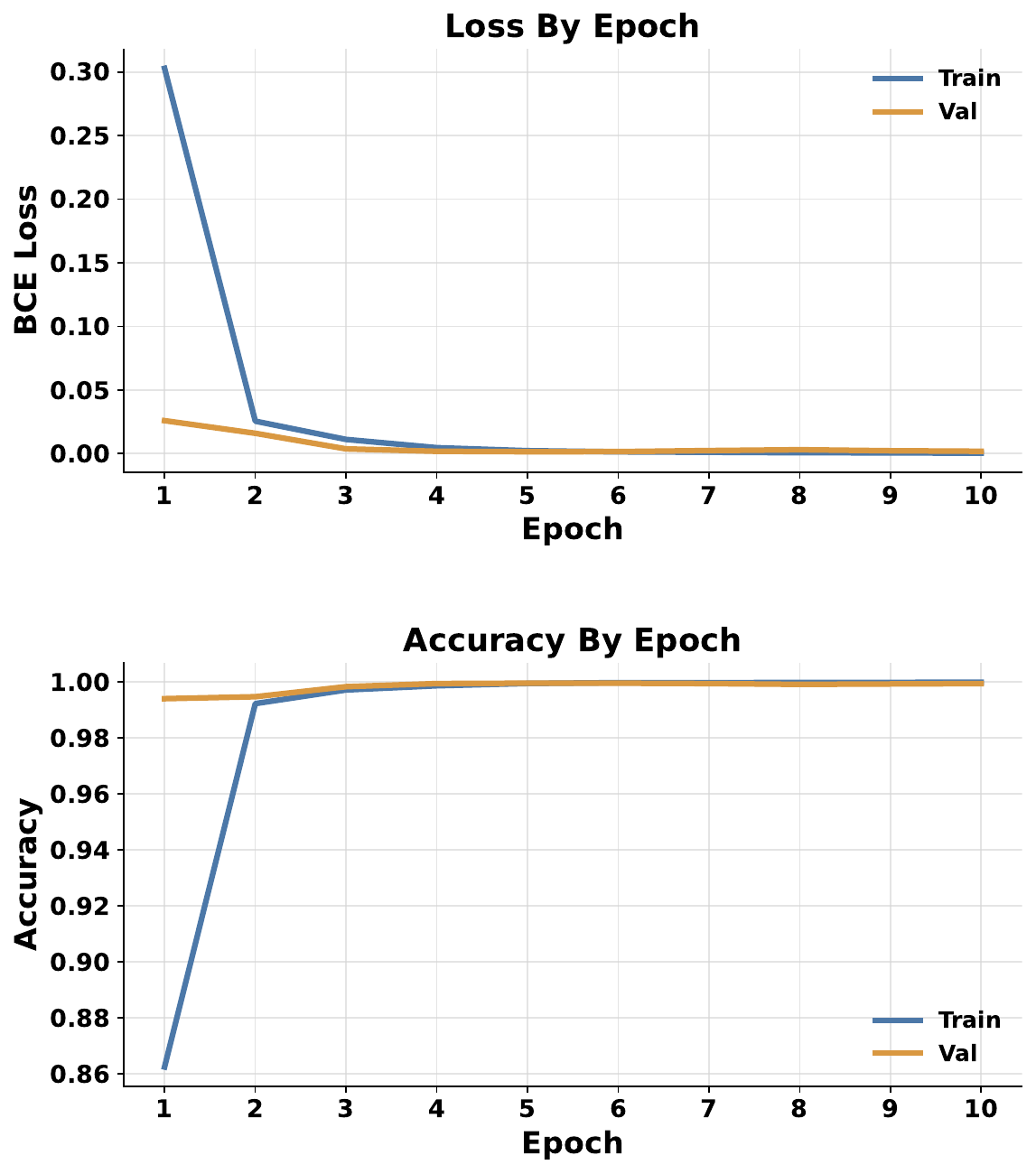} 
   \caption{(top) Cross-entropy loss as a function of epoch for a typical training trial. Loss on training set in blue, loss on validation set in orange. (bottom) Classification accuracy on training (blue) and validation (orange) sets as a function of epoch.}
   \label{fig2}
\end{figure}

\section{Experimental Design}

We split the 36,171 unique lower-case 9-mers corresponding to viable avidian
replicators into training, validation, and test sets using an 80/10/10 split.
To construct a balanced binary classification task, we sampled one non-replicator 9-mer for each positive sequence (replicator) in each split, drawing uniformly at random from the full space of 9-mers while excluding all known replicators. This produced balanced data sets containing 57,872 training sequences, 7,234 validation sequences, and 7,236 test sequences. Our focal classifier is a multi-layer perceptron (MLP) with a 32-dimensional character embedding, followed by hidden layers of sizes 512 and 256 with Gaussian Error Linear Unit (GELU) activations and dropout of 0.1. We optimize the model using AdamW with learning rate \(2\times10^{-3}\) and weight decay \(10^{-4}\), using a batch size of 8,192 for 10 epochs. To test the robustness of the MLP design, we also varied the embedding, the layer structure, as well as the activation function.

\begin{table}[t]
\centering
\caption{Spoofing Experiment Design}
\label{tab:spoof_design}
\begin{tabular}{lr}
\hline
Component & Value \\
\hline
Replicate count & 30 \\
Uniform starts per seed (a--z) & 26 \\
Random starts per seed & 26 \\
Total runs per seed & 52 \\
Total uniform runs & 780 \\
Total random runs & 780 \\
Total spoof runs & 1560 \\
Target class & Replicator \\
Query budget & 300 \\
Sequence length & 9 \\
Ground truth & AVIDA Replicator 9-mers \\
\hline
\end{tabular}
\end{table}

Fig.~\ref{fig2} shows the evolution of the Binary Cross Entropy (BCE, a loss function) as a function of epoch for the training and validation set, demonstrating that our MLP can classify avidian sequences near perfectly: on the balanced test split (50/50 viable and non-viable sequences), the model achieved 99.97\% accuracy with a BCE loss of 0.00088. Using a 0.5 decision threshold produced 3,618 true positives, 3616 true negatives, 2 false positives and no false negatives, corresponding to a precision of 99.94\% and  100.00\% recall. The train/validation curves show rapid convergence within a few epochs and near-saturation thereafter. This indicates that the local decision problem
on the sampled balanced distribution is easy for the MLP, at least under the chosen negative-sampling regime.  Standard supervised metrics alone suggest the model has essentially solved the task. However, balanced test accuracy does not probe how confidence behaves far from the empirical data manifold. Because true replicators are exponentially sparse in the full $26^9$ space, even a tiny mismatch between the learned confidence surface and the true replicator manifold can create a large volume of high-confidence false positives. The experiments below test exactly this failure mode. 

To generate ``spoofed" sequences (``spoofing" refers to the process we now describe), we performed a greedy hill-climbing search in the space of 9-mers. Starting from either one of the 26 uniform sequences ({\tt aaaaaaaaa} to {\tt zzzzzzzzz}) or a randomly sampled 9-mer, we repeatedly proposed a single-site mutation and evaluated the classifier's confidence that the mutated sequence was a replicator. If the mutation increased that confidence, we accepted it; otherwise, we rejected it. Thus, the search climbed the model's confidence landscape toward sequences predicted with increasingly high confidence to be replicators.

\begin{figure}[htbp] 
   \centering
   \includegraphics[width=3.2in]{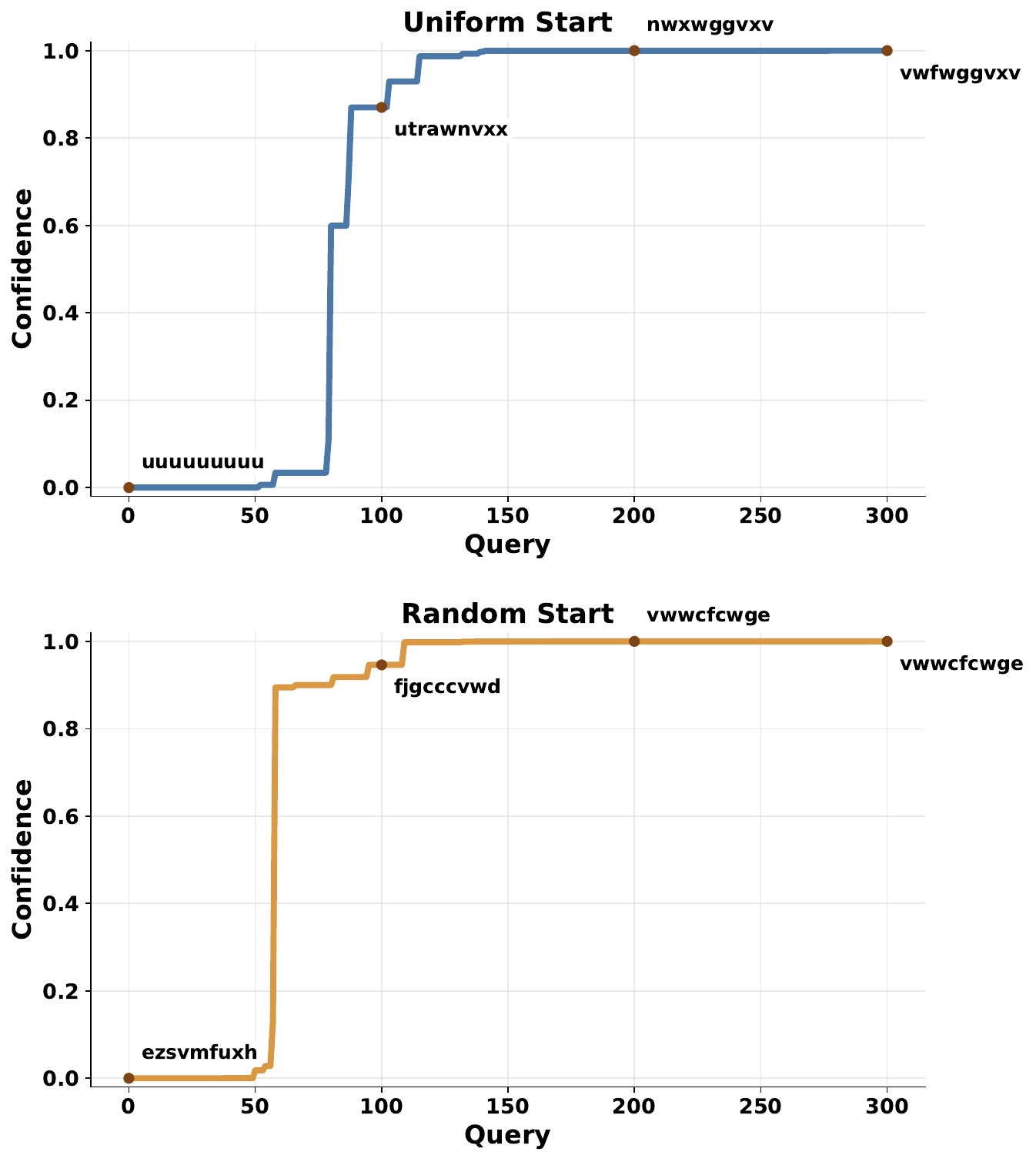} 
   \caption{Representative string evolution for a single run using the MLP architecture: Confidence of the classifier for an evolved 9-mer being a replicator as a function of iteration of the greedy guided walk, for a run starting with a uniform (above) / random (below) initial string. Note that \textbf{none} of the annotated 9-mers along the string evolution are replicators.}
   \label{fig3}
\end{figure}

Formally, if \(p(\mathbf{s})\) denotes the classifier's predicted probability that sequence \(\mathbf{s}\) is a replicator, then at each step we accepted a proposed mutant \(\mathbf{s}'\) only when
\begin{equation}
p(\mathbf{s}') > p(\mathbf{s}_t),
\label{eq1}
\end{equation}
where \(\mathbf{s}_t\) is the current sequence. The search continues until the pre-specified query budget was exhausted. Fig.~\ref{fig3} shows a typical spoofing trajectory, with model confidence as a function of iteration. To test the robustness of this search procedure, we also ran supplementary experiments with a random walk (no selection), a stochastic random walk (probabilistic acceptance of a move), as well as two evolutionary methods, one using tournament selection and one using truncation selection (accept top 20\%).

\section{Results}
\begin{figure}[htbp] 
   \centering
   \includegraphics[width=3in]{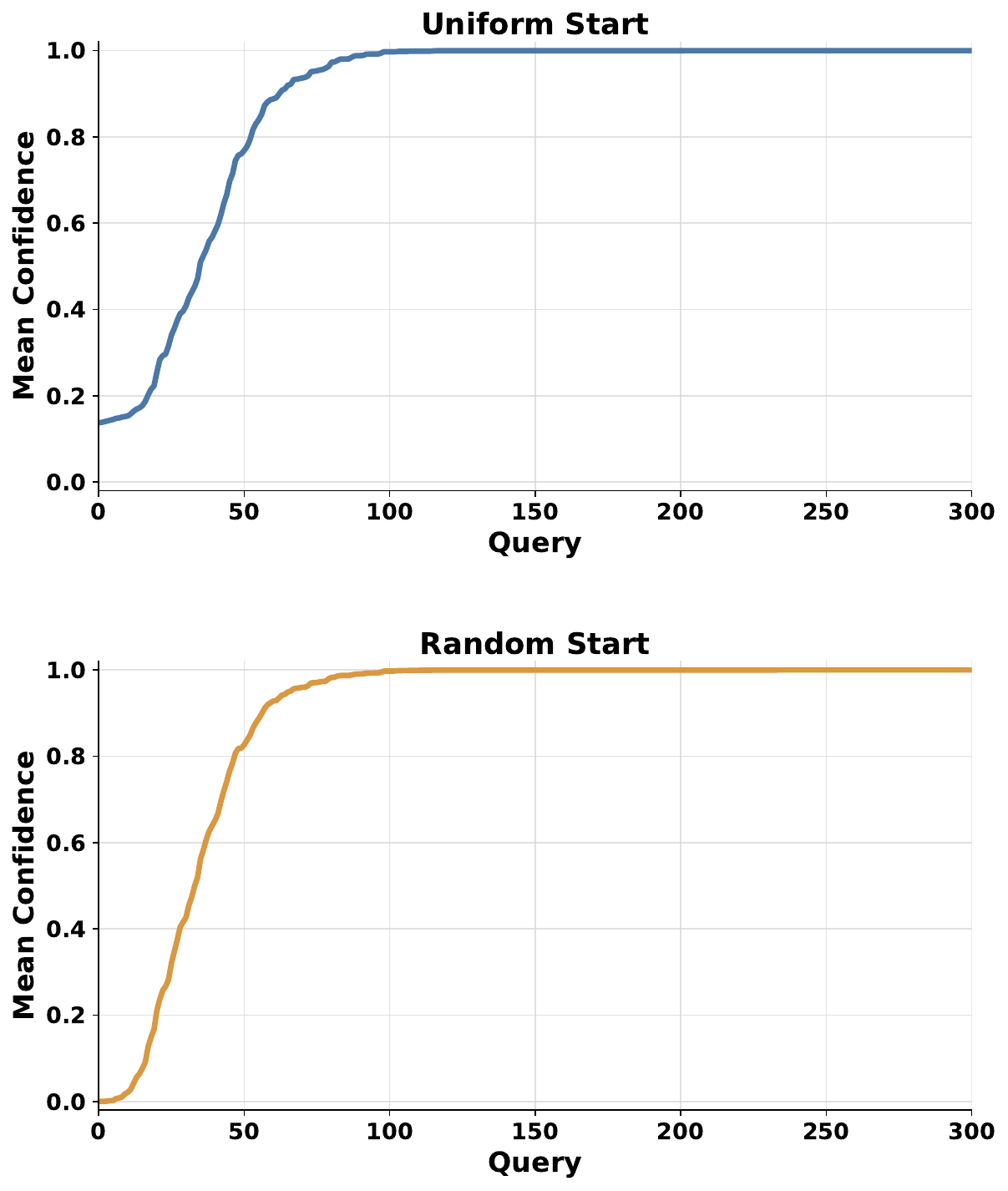} 
   \caption{Mean spoofing confidence over model queries from 30 replicates, for two different initial conditions: uniform sequences (polymers of all {\tt a} to all {\tt z}) and random sequences.}
   \label{fig4}
\end{figure}

For each of the test conditions (fixed AI classifier, search method, and MLP architecture) we ran a total of 1560 (52x30) independent spoofing runs, half starting with the 26 uniform letter sequences and the other half with 26 random letter sequences\footnote{In the following, we will discuss the results of the MLP architecture, while showing results obtained with variations in the Supplementary Experiments section.}.
Of these, we found all runs resulting in 100\% spoofing confidence as early as 150 model queries. Also, as seen in Tab.~\ref{tab:confidence_by_query}, we observed significant spoofing confidence of 82.66\% and 76.85\%, as early as model query count of 50, for random and uniform starting sequences respectively. Fig.~\ref{fig3} shows a typical spoofing trajectory, showing model confidence as a function of query count. We also show an aggregate of spoofing confidence over 30 replicates in Fig.~\ref{fig4}

\begin{table}[ht]
\centering
\caption{Mean spoofing confidence of evolving 9-mers across model query count.}
\label{tab:confidence_by_query}
\begin{tabular}{rcc}
\hline
Query & Random Start & Uniform Start \\
\hline
0   & 0.0000 & 0.1375 \\
25  & 0.3203 & 0.3409 \\
50  & 0.8266 & 0.7685 \\
75  & 0.9708 & 0.9532 \\
100 & 0.9972 & 0.9973 \\
125 & 0.9998 & 0.9997 \\
150 & 1.0000 & 1.0000 \\
175 & 1.0000 & 1.0000 \\
200 & 1.0000 & 1.0000 \\
\hline
\end{tabular}
\end{table}
\begin{figure}[htbp] 
   \centering
   \includegraphics[width=2.75in]{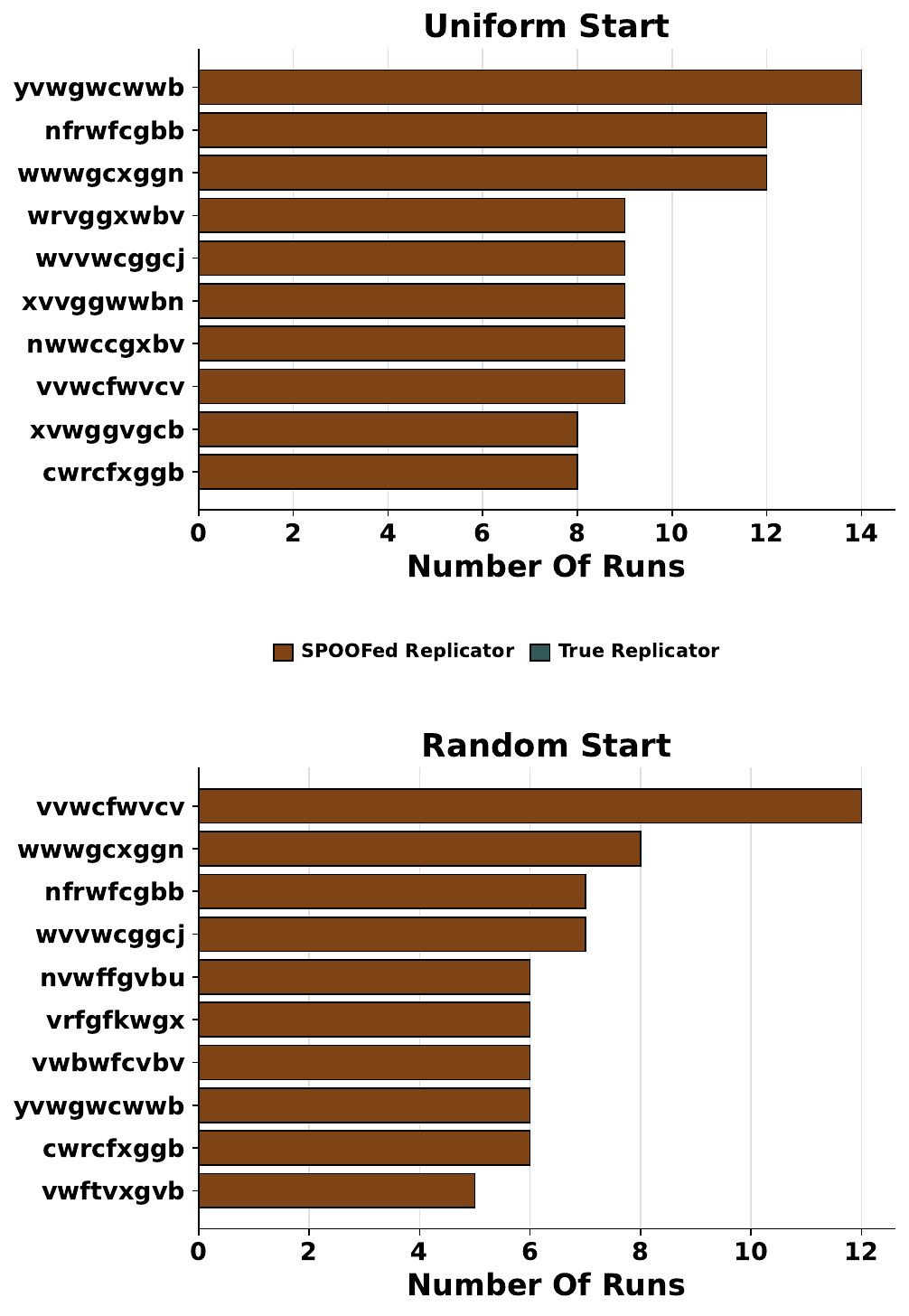} 
   \caption{Frequency of final 9-mers evolved from 780 spoofing runs (from 30 replicates of 26 initial starting 9-mers), for uniform (top) and random (bottom) starts.}
   \label{fig5}
\end{figure}
It is clear that the simple optimizer (optimizing Eq.~(\ref{eq1}) via a greedy hill climb) finds sequences that the model considers to be replicators with certainty, while those optima are not, in fact, actual replicators. 

The spoof budget was modest as we are using a mere 300 model queries, yet actually achieve perfect spoofing confidences as early as 150 model queries. This means the failure is not a pathological consequence of an enormous search: a few hundred model queries are enough to climb into highly-confident false basins. 

\begin{figure}[htbp] 
   \centering
   \includegraphics[width=3in]{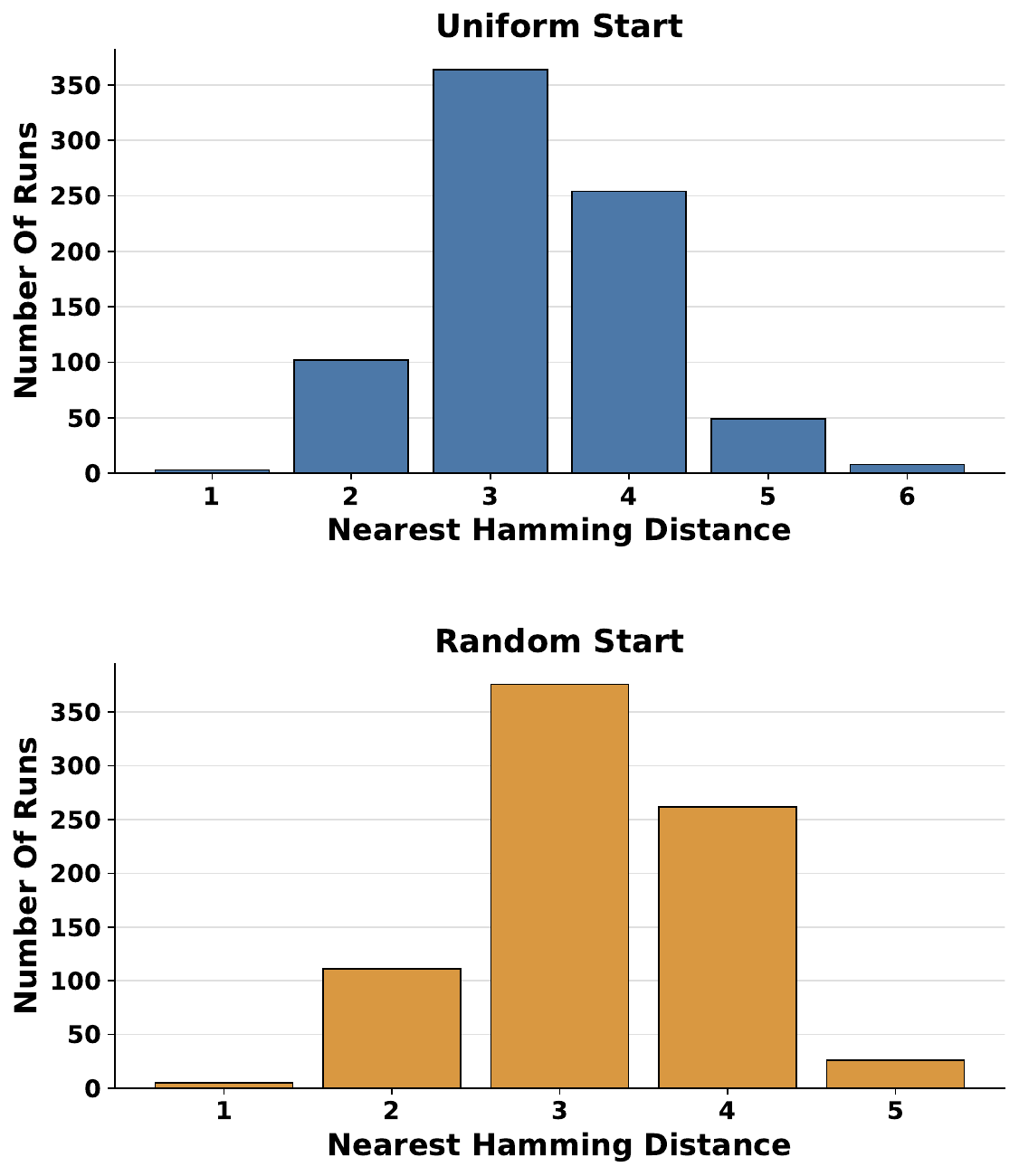} 
   \caption{Hamming distance of final 9-mers evolved from 780 spoofing runs (from 30 replicates of 26 initial starting 9-mer), for uniform (top) and random (bottom) starts.}
   \label{fig5a}
\end{figure}

The 9-mer endpoints that emerged at the end of the spoofing runs are not random.
Across 1,560 total runs (780 from uniform starts and 780 from random starts),
the search repeatedly converged to a small set of terminal strings, indicating
that the confidence landscape contains a limited number of strong attractors.
For uniform starts, the most common endpoint was {\tt yvwgwcwwb}, which
appeared 14 times, followed by {\tt nfrwfcgbb} and {\tt wwwgcxggn}, each of
which appeared 12 times. For random starts, the most common endpoint was
{\tt vvwcfwvcv}, which appeared 12 times, followed by {\tt wwwgcxggn}, which
appeared 8 times (Fig.~\ref{fig5}). The Hamming-distance distributions in
Fig.~\ref{fig5a} further show that the final sequences are typically only a few
mutations away from the nearest true replicator, with most runs ending at
distance 3 or 4. Together, these results suggest that the model has learned a
coarse replicator-like motif family without capturing the exact combinatorial
constraints that distinguish true self-replicators from nearby impostors.
\begin{figure}[htbp] 
   \centering
   \includegraphics[width=3.5in]{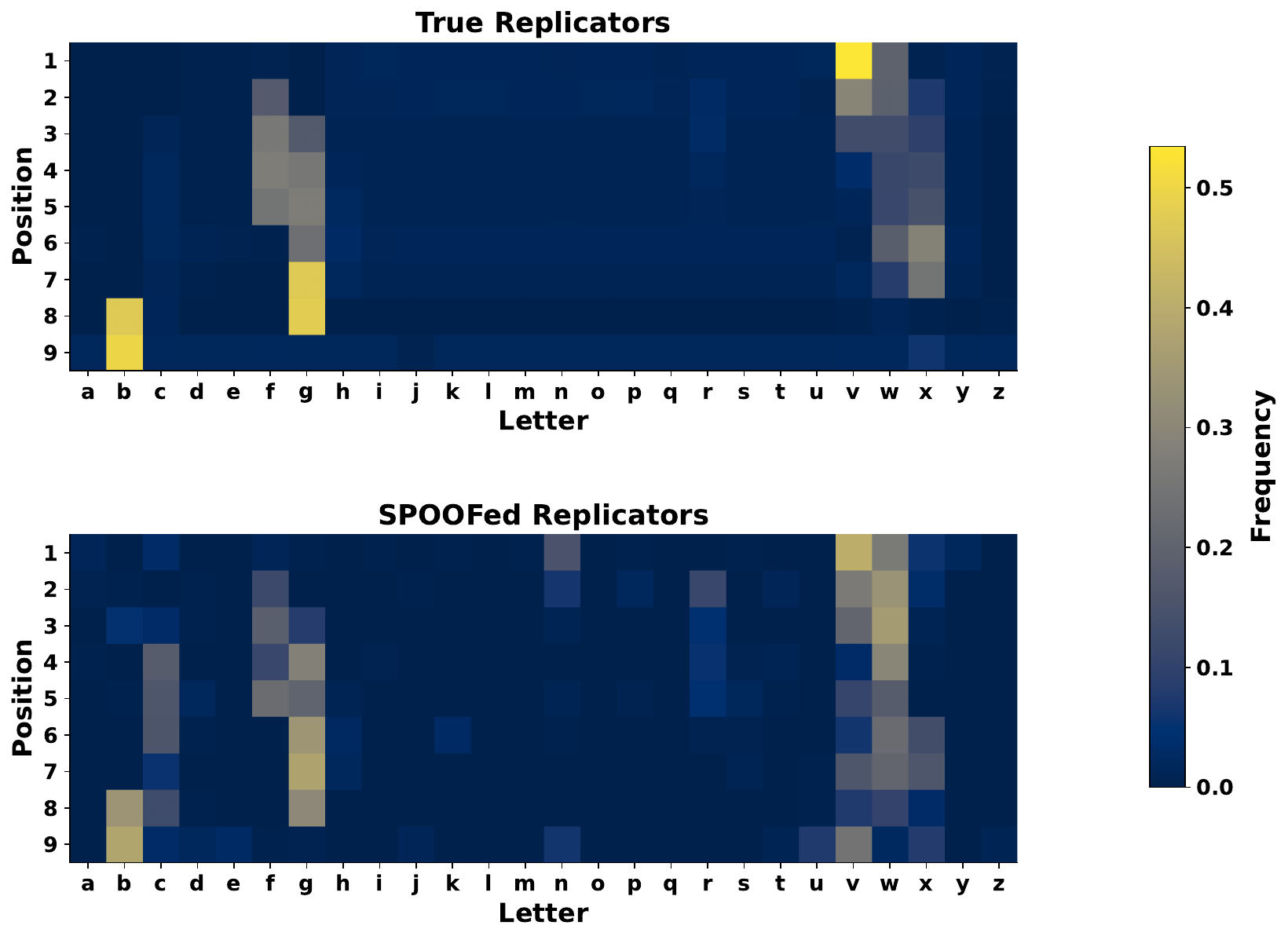} 
   \caption{Frequency of symbol (color bar on the right) as a function of position for true replicators (above) and false (spoofed) replicator evolved (below).}
   \label{fig6}
\end{figure}

The position-wise heat maps (Fig.~\ref{fig6}) reinforce that point: spoof endpoints concentrate on a narrow motif pattern that resembles the empirical replicator distribution without matching it exactly. The motif patterns seen in the true replicators in Fig.~\ref{fig6} are precisely those that are used to distinguish the two different types of replicators observed in the length-8 landscape already~\citep{Nitashetal2017}.
Varying key elements of the procedure, such as the classifier being used, the architecture of the MLP, as well as the search method, does not change the conclusions (see section Supplementary Experiments).

\section{Discussion}
We find that a simple multi-layer perceptron, as well as any other of a number of alternative methods, can achieve near-perfect held-out accuracy on a balanced sampled test data set while still being catastrophically vulnerable to targeted confidence maximization in sequence space. The spoof procedure reliably discovers high-confidence attractors, all of them false positives. This means the learned score function is useful as a discriminative classifier on the sampled train/test distribution, yet not reliable as a generative or optimization oracle over the full combinatorial domain.
If this model were used to guide a search for novel replicators, naive hill-climbing on model confidence would spend most of its effort in deceptive basins. 

The fact that false attractors are typically only a few mutations away
from true replicators is scientifically interesting: it points to a
near-replicator ``shell" of sequences where local motif statistics are strong but exact
functional constraints are still violated. In our spoofing experiments,
the Hamming-distance distributions are concentrated primarily at distances
3 and 4 from the nearest true replicator, for both uniform and random
starting conditions (Fig.~\ref{fig5a}), which is a significant percentage of the total number of features. 
In larger landscapes, we expect the false fixed points to be significantly  
\begin{table*}[!t]
\centering
\caption{
MLP architecture ablation after 300 model queries. The training data,
training protocol, and greedy hill-climbing search are fixed; only
embedding dimension, hidden-layer widths, and activation function are
varied. ``Mean confidence'' is reported as a percentage and denotes the
mean best classifier confidence assigned to searched sequences within the
query budget; in this setting, it measures fooling confidence rather than
replicator discovery. ``True replicators'' is the percentage of runs in
which the best sequence encountered by query 300 was a known Avida
length-9 replicator.
}
\label{tab:mlp_architecture_ablation}
\setlength{\tabcolsep}{1pt}
\renewcommand{\arraystretch}{1.08}
\begin{tabular}{lccc c cc cc}
\toprule
& \multicolumn{3}{c}{Architecture} & &
\multicolumn{2}{c}{Uniform Starts} &
\multicolumn{2}{c}{Random Starts} \\
\cmidrule(lr){2-4} \cmidrule(lr){6-7} \cmidrule(lr){8-9}
Condition
& Embedding
& \makecell{Hidden\\Layers}
& Activation
& \makecell{Test\\Accuracy (\%)}
& \makecell{Mean\\Confidence (\%)}
& \makecell{True\\Replicators (\%)}
& \makecell{Mean\\Confidence (\%)}
& \makecell{True\\Replicators (\%)} \\
\midrule
Reference MLP & 32 & 512, 256  & GELU & 99.96 & 100.00 & 0.00 & 100.00 & 0.00 \\
Matched MLP   & 32 & 512, 256  & ReLU & 99.93 & 100.00 & 0.13 & 100.00 & 0.26 \\
Small MLP     & 16 & 128, 64   & GELU & 99.94 & 100.00 & 2.82 & 100.00 & 3.85 \\
Large MLP     & 64 & 1024, 512 & GELU & 99.96 & 92.31  & 0.00 & 100.00 & 0.00 \\
\bottomrule
\end{tabular}
\end{table*}
distant from the true ones. 
By way of example, the number of features in the Random Forest model\footnote{Note that {\em all} machine learning methods that rely on fitting a probability distribution---and that we have tested so far---are susceptible to spoofing.} applied to py-GC-MS data is 8,149~\citep{Wongetal2025}. The fixed points of spoofed images from ImageNet (color images with 150,528 features), classified by the current state-of-the-art Deep Neural Networks and transformers, were completely unrecognizable~\citep{Guptaetal2025}.  Furthermore, we expect in larger landscapes the number of false fixed points to dwarf the number of true ones.

The supplementary experiments show that the described phenomenon is not unique to multi-layer perceptrons: we tested convolutional neural networks (CNNs), transformers, SVMs (Support Vector Machines), Random Forests, XGB (eXtreme Gradient Boosting), and even simple logistic regression. Random Forests showed minor resistance relative to other classifiers while remaining significantly susceptible to spoofing. We can therefore confidently state that this property (vulnerability to out-of-distribution samples) is shared by all machine learning methods. 

How representative are these experiments with Artificial Life when compared with state-of-the-art efforts to infer the presence of life in chemical signatures? Typical methods will analyze chemical signatures (for example, in the form of mass spectrometry spectra) rather than molecular sequences. From this point of view, the present study can only suggest a vulnerability, but certainly does not indicate it. A future investigation attempting to spoof the actual pyrolysis-gas chromatography-mass spectrometry (py-GC-MS) data would provide more insight. 

We conclude that if false positives (false high-confidence fixed points) outnumber true positives  in samples from extraterrestrial measurements (because they are outside of the distribution that an AI was trained on), we risk accepting high-confidence classifications at
face value. If the proven vulnerability of AI methods to out-of-distribution high-confidence failures translates to AI-informed search for life, using such methods on space missions has a high probability of undermining public confidence in Astrobiology missions.

\section{Supplementary Experiments}
In order to test whether the vulnerability to high-confidence misclassification is due to a particular choice of experimental conditions, we perform ``ablation" experiments in which we change one of the design elements while keeping all others the same. In the following, we test the robustness of the MLP architecture, replace the MLP with six alternative AI methods, and replace the search algorithm.  

Unless otherwise noted, each ablation follows the same run structure as the
main experiments: 1,560 spoofing runs, with 30 random seeds $\times$ 52 starts,
26 uniform starts and 26 random starts per seed, and a 300-query budget per run.
For all ablation tables, mean confidence denotes the mean best
classifier confidence encountered up to a given query count:
that is, for each run we record the largest value of $p(s)$ seen so far,
where $p(s)$ is the classifier's predicted probability that sequence
$s$ is a replicator, and then average this value over runs. Thus,
high mean confidence indicates that the classifier can be fooled
with high confidence; it does not imply that the searched sequences
are true replicators. Tables report the same best-seen statistic at
the end of the 300-query budget.

\subsection{MLP Architecture Ablation}
\begin{table*}[t]
\centering
\caption{
Classifier ablation after 300 model queries. Greedy hill climbing is
fixed while the classifier is varied. ``Mean confidence'' is reported as
a percentage and denotes the mean best classifier confidence assigned to
searched sequences within the query budget; in this setting, it measures
fooling confidence rather than replicator discovery. ``True replicators''
is the percentage of runs in which the best sequence encountered by query
300 was a known Avida length-9 replicator.
}
\label{tab:model_ablation}
\setlength{\tabcolsep}{9pt}
\renewcommand{\arraystretch}{1.08}
\begin{tabular}{l c cc cc}
\toprule
& & \multicolumn{2}{c}{Uniform Starts} & \multicolumn{2}{c}{Random Starts} \\
\cmidrule(lr){3-4} \cmidrule(lr){5-6}
Model
& \makecell{Test\\Accuracy (\%)}
& \makecell{Mean\\Confidence (\%)}
& \makecell{True\\Replicators (\%)}
& \makecell{Mean\\Confidence (\%)}
& \makecell{True\\Replicators (\%)} \\
\midrule
Reference MLP       & 99.96 & 100.00 & 0.00  & 100.00 & 0.00 \\
Logistic Regression & 99.96 & 100.00 & 0.00  & 100.00 & 0.00 \\
SVM                 & 99.90 & 100.00 & 0.00  & 100.00 & 0.00 \\
Random Forest       & 99.90 & 40.76  & 12.69 & 74.53  & 26.03 \\
XGB                 & 99.94 & 96.64  & 3.08  & 96.64  & 3.59 \\
CNN                 & 99.97 & 76.55  & 0.51  & 100.00 & 1.03 \\
Transformer         & 99.90 & 100.00 & 0.00  & 100.00 & 0.00 \\
\bottomrule
\end{tabular}
\end{table*}

\begin{table*}[t]
\centering
\caption{
Search ablation after 300 model queries. The reference MLP classifier is
fixed while the search procedure is varied. ``Mean confidence'' is reported
as a percentage and denotes the mean best classifier confidence assigned to
searched sequences within the query budget; in this setting, it measures
fooling confidence rather than replicator discovery. ``True replicators''
is the percentage of runs in which the best sequence encountered by query
300 was a known Avida length-9 replicator.
}
\label{tab:search_ablation}
\setlength{\tabcolsep}{15pt}
\renewcommand{\arraystretch}{1.08}
\begin{tabular}{l cc cc}
\toprule
& \multicolumn{2}{c}{Uniform Starts} & \multicolumn{2}{c}{Random Starts} \\
\cmidrule(lr){2-3} \cmidrule(lr){4-5}
Search Condition
& \makecell{Mean\\Confidence (\%)}
& \makecell{True\\Replicators (\%)}
& \makecell{Mean\\Confidence (\%)}
& \makecell{True\\Replicators (\%)} \\
\midrule
Greedy Hill Climb     & 100.00 & 0.00 & 100.00 & 0.00 \\
Random Walk           & 28.70  & 0.00 & 19.30  & 0.00 \\
Stochastic Hill Climb & 47.00  & 0.00 & 43.09  & 0.00 \\
Tournament Selection  & 75.80  & 0.00 & 87.27  & 0.00 \\
Truncation Selection  & 98.32  & 0.00 & 99.26  & 0.00 \\
\bottomrule
\end{tabular}
\end{table*}

We first test whether spoofing depends on the particular architecture
of the reference MLP. The training data, training protocol, and greedy
hill-climbing search are fixed, while the MLP embedding dimension,
hidden-layer widths, and activation function are varied. The reference
MLP maps each Avida instruction to a learned 32-dimensional embedding,
flattens the embedded length-9 sequence, and passes it through hidden
layers of widths 512 and 256 with GELU activations. We compare this
reference model to a same-width ReLU variant, a smaller GELU MLP, and
a larger GELU MLP.

Table~\ref{tab:mlp_architecture_ablation} 
shows that high-confidence
spoofing is robust to these architectural changes. Most variants reach
mean confidence near $1.0$ within 300 model queries from both uniform
and random starts. The large MLP is somewhat less susceptible from
uniform starts, but still reaches perfect mean confidence from random
starts. True-replicator recovery remains rare, showing that high
classifier confidence is usually achieved by non-replicating sequences
rather than by discovery of actual Avida replicators.

\subsection{Classifier Ablation}

We next test whether spoofing is specific to the reference MLP. The
search procedure is fixed to greedy hill climbing, while the classifier
is varied. Logistic regression and the linear SVM use one-hot encodings
of the length-9 sequence. The Random Forest and XGB models are tree
ensembles trained on the same one-hot representation, with XGB using
gradient-boosted trees. The CNN uses a learned 32-dimensional instruction
embedding, two one-dimensional convolutional layers with 64 and 128
channels, max pooling over sequence positions, and a dense prediction
head. The small transformer encoder uses a learned 32-dimensional
instruction embedding, positional embeddings, two self-attention encoder
layers, four attention heads, and a dense prediction head.

Table~\ref{tab:model_ablation} 
shows that high-confidence spoofing is not specific to the reference MLP.
Logistic regression, the SVM, and the transformer rapidly reach mean
confidence near $1.0$ from both start distributions. XGB also reaches
high confidence, while the Random Forest and CNN show slower or lower
confidence trajectories in some settings. The Random Forest recovers true
replicators more often than the other classifiers, but its confidence
trajectory still demonstrates that high held-out accuracy alone is not a
sufficient guarantee of reliable behavior under confidence-directed search.
Overall, the classifier ablation shows that the failure mode is not tied
to one neural architecture.

\subsection{Search Ablation}
Finally, we test whether spoofing depends on the search procedure. The classifier is fixed to the reference MLP, while the search method is varied. Greedy hill climbing accepts only confidence-increasing single-character mutations. Random walk uses the same mutation operator but accepts every mutation, serving as a no-selection control. Stochastic hill climbing uses fixed-temperature Metropolis acceptance, allowing occasional lower-confidence moves. The population methods maintain 30 candidate sequences for 10 generations, using either tournament selection or truncation selection from the highest-confidence candidates.

Table~\ref{tab:search_ablation} shows that search pressure strongly controls the rate and extent of spoofing. Greedy hill climbing reaches near-perfect mean confidence fastest, random walk remains low, and stochastic hill climbing lies between them. The population searches, especially truncation selection, recover much of the greedy-search behavior. Across all search procedures, however, high spoofing confidence does not correspond to reliable recovery of true replicators.

\section{Acknowledgements}
This research was not funded. We gratefully acknowledge the support of the Michigan State University High Performance Computing Center and the Institute for Cyber Enabled Research (iCER).


\end{document}